\documentclass[10pt,journal,compsoc]{IEEEtran}
\usepackage{amsmath,amssymb,amsfonts}
\usepackage{algorithmic}
\usepackage{graphicx}
\usepackage{textcomp}
\usepackage{subfigure}
\usepackage{multirow}
\usepackage{color}
\usepackage[colorlinks,linkcolor=blue]{hyperref}

\ifCLASSOPTIONcompsoc
  \usepackage[nocompress]{cite}
\else
  \usepackage{cite}
\fi
\ifCLASSINFOpdf
\else
\fi
\begin{document}
%
\title{A Similarity-based Framework \\for Classification Task}
%
%
%
%

\author{Zhongchen Ma, and Songcan Chen
\IEEEcompsocitemizethanks{\IEEEcompsocthanksitem Corresponding author: Songcan Chen. The authors are with the College of Computer Science and Technology, Nanjing University of Aeronautics and Astronautics, Nanjing 211106, China (e-mail: zhongchen.ma@nuaa.edu.cn; s.chen@nuaa.edu.cn).}
\thanks{Manuscript received July 23, 2018.}}

\author{Zhongchen Ma, and Songcan Chen
	\IEEEcompsocitemizethanks{\IEEEcompsocthanksitem Zhongchen Ma is with both the School of Computer Science \& communications Engineering, Jiangsu University,  and  the College of Computer Science and Technology, Nanjing University of Aeronautics and Astronautics, Zhenjiang 212013, China (e-mail: zhongchen.ma@nuaa.edu.cn).}
	\IEEEcompsocitemizethanks{\IEEEcompsocthanksitem Corresponding author: Songcan Chen is with the College of Computer Science and Technology, Nanjing University of Aeronautics and Astronautics, and the MIIT Key Laboratory of Pattern Analysis and Machine Intelligence, Nanjing 211106, China (e-mail: s.chen@nuaa.edu.cn).}
	\thanks{Manuscript received July 23, 2018.}}

\markboth{Journal of \LaTeX\ Class Files,~Vol.~14, No.~8, August~2015}%
{Shell \MakeLowercase{\textit{et al.}}: Bare Demo of IEEEtran.cls for Computer Society Journals}
%



\IEEEtitleabstractindextext{%
\begin{abstract}
Similarity-based method gives rise to a new class of methods for multi-label learning and also achieves promising performance. In this paper, we generalize this method, resulting in a new framework for classification task. Specifically, we unite similarity-based learning and generalized linear models to achieve the best of both worlds. This allows us to capture interdependencies between classes and prevent from impairing performance of noisy classes. Each learned parameter of the model can reveal the contribution of one class to another, providing interpretability to some extent. Experiment results show the effectiveness of the proposed approach on multi-class and multi-label datasets.
\end{abstract}

}

\maketitle

\IEEEdisplaynontitleabstractindextext

%
\IEEEpeerreviewmaketitle

\IEEEraisesectionheading{\section{Introduction}\label{sec:introduction}}
Multi-class classification (MCC) and multi-label classification (MLC) are two typical paradigms in machine learning and can be widely applicable in such fields as computer vision, text classification, medical image, information retrieval, etc. Traditionally, MCC is the problem of classifying instances into one of three or more classes, while MLC is the problem of allowing instances to belong to several classes (labels) simultaneously. Therefore, MCC can be cast into MLC by restricting each instance to have only one label.

Distance-based classifiers, which refer to a class of classifiers basing its decision on the distance calculated from the target instance to the training instances, are simple but effective methods for MCC. Based on what type of distance information is used, such methods can be further categorized as instance-information-based and class-information-based classifiers. The instance-information-based classifiers classify an instance based on the distance information in the feature space, while class-information-based classifiers classify an instance based on the distance information in the space induced by classes. In particular, $k$-nearest neighbor (KNN) classifier \cite{altman1992introduction} and the nearest class mean (NCM) classifier \cite{schutze2008introduction} are representative instance-information-based and class-information-based methods, respectively. Specifically, KNN assumes that similar instances in the feature space very likely belongs to the same class, thus classifying an instance by a majority vote of its $k$-nearest neighbors, and in another way, NCM represents a class by the mean feature vector of its instances, thus assigning an instance to the class with the closest mean.

As a generalization of MCC, MLC inevitably brings more difficulties to learning. To address these challenges of MLC, many methods have been derived from the above distance-based classifiers \cite{zhang2014review}. For example, multi-label $k$-nearest neighbor (MLKNN) \cite{zhang2007ml} and instance-based learning by logistic regression (IBLR) \cite{cheng2009combining} are two effective KNN-based multi-label methods. Based on the type of distance information used, both methods can be classified as instance-information-based classifiers.

Compared to class-information-based classifiers, instance-information-based classifiers more likely misclassify an instance in the class boundary, because the assumption of such a type of classifiers that similar instances in the feature space very likely belong to the same class, is more likely violated in that case. However, contrary to extensively used instance-information-based classifiers for MCC and MLC, there are currently few class-information-based classifiers for both tasks. Recently, a similarity-based  (analogous to distance-based) multi-label learning (SML) classifier\cite{rossi2018similarity} was proposed, which represents a label by all its instances and assigns an instance to the labels with large similarities, thus belonging to class-information-based classifiers. It forms a new class of methods for MLC with promising performance\cite{rossi2018similarity}. However, we witness: {1) SML is a lazy learning method and often deals much worse with noise in the training data than inductive learning methods (e.g., generalized linear models) \cite{kotsiantis2007supervised}; 2) SML does not take correlations and interdependencies between labels into account, its potentials have not yet been fully exploited. 

To remedy the above problems, we try to construct a learning framework, which is able to: 1) achieve the best of SML and generalized linear models, 2) capture interdependencies between classes. 
Specifically, this framework unifies \emph{lazy} SML and \emph{inductive} generalized linear models to achieve the best of both worlds, as opposed to IBLR \cite{cheng2009combining} that unifies \emph{lazy} KNN and \emph{inductive} logistic regression model. A key idea of our approach is to consider the similarities between the query $\mathbf{x}_0$ and the instance sets of different classes as the input features to the generalized linear models, and then to optimize this model on training data to deduce an inductive optimal classifier. Intuitively, the so-learned parameters of the model can reflect the contribution of one class to another class, providing interpretability to some extent. Moreover, by using an $\ell_1$-norm penalty to sparsify the solution of the model, the proposed approach is further also able to discover irrelevant classes, this way both preventing from impairing performance of them and providing better interpretability.

The rest of this paper is organized as follows: The problems of MLC and MCC are introduced in a more formal way in Sect. 2, and our novel method is then described in Sect. 3. Section 4 is devoted to experiments on MLC and MCC datasets. The paper ends with a summary and some concluding remarks in Sect. 5.

\section{MCC and MLC}
Let $\mathbb{X} \subset \mathbb{R}^d$ denote an instance space and let $\mathcal{L}=\{\lambda_1, \lambda_2, \cdots, \lambda_m\}$ ($m \geq 3$) be a finite set of class labels. Moreover, we are given a training set $\mathcal{D}=\{(\mathbf{x}_i, L_i)\}_{i=1}^N$, where $N$ is the number of instances and $\mathbf{x}_i \in \mathbb{X}$. Based on the label association to the instances, the classification problem can be categorized into MCC and MLC.
Formally, for $\forall i \in \{1, 2, \cdots, N\}$,

1) if $L_i$ is an element of $\mathcal{L}$, the classification problem is MCC;

2) if $L_i$ is a subset of $\mathcal{L}$, the classification problem is MLC. This subset is often called the set of relevant labels, while the complement $\mathcal{L}\setminus L_i$ is considered as irrelevant for $\mathbf{x}_i$.

Both MLC and MCC have received a great deal of attention in machine learning. And a number of methods have been developed, among which the distance-based classifiers are considered as simple yet effective method. Recently, a new distance-based classifier, SML \cite{rossi2018similarity} was proposed with promising performance. However, when applied to MLC and MCC, as stated in the introduction section, it has some key shortcomings needing to be addressed. Thus, in this paper, we are especially interested in SML to improve its performance in MLC and MCC tasks.

\section{Similarity-based Learning}

\subsection{SML classifier}
Let us assume that all the given feature vectors are normalized from its original range to the range $[-1, 1]$ and define the subset $\mathcal{D}_k \subseteq \mathcal{D}$ of training instances with class $\lambda_k \in \mathcal{L}$ as
\begin{equation}
\mathcal{D}_k=\{(\mathbf{x}_i, L_i) \in \mathcal{D} | \lambda_k \in L_i\},
\end{equation}
then, the weight $f_k(\mathbf{x}_i)$ of class $\lambda_k$ for an unseen instance $\mathbf{x}_i \in \mathbb{X}$ is estimated as:
\begin{equation}
f_k(\mathbf{x}_i)=\sum_{\mathbf{x}_j \in \mathcal{D}_k} \Phi \langle \mathbf{x}_i, \mathbf{x}_j\rangle
\end{equation}
where $\Phi$ is an arbitrary similarity function. Then, all class weights denoted by $f(\mathbf{x}_i)$ are estimated as:
\begin{equation}
f(\mathbf{x}_i)=\left[\begin{array}{c}
f_1(\mathbf{x}_i)  \\
\vdots \\
f_m(\mathbf{x}_i) \end{array} \right]=\left[\begin{array}{c}
\sum_{\mathbf{x}_j \in \mathcal{D}_1} \Phi \langle \mathbf{x}_i, \mathbf{x}_j\rangle\\
\vdots\\
\sum_{\mathbf{x}_j \in \mathcal{D}_m} \Phi \langle \mathbf{x}_i, \mathbf{x}_j\rangle
\end{array} \right]
\end{equation}
After estimating $f(\mathbf{x}_i)=[f_1(\mathbf{x}_i), \cdots, f_m(\mathbf{x}_i)]^T \in \mathbb{R}^m$, the class set $L_i$ of $\mathbf{x}_i$ can be predicted as the classes with $\xi_i$ largest values of $f(\mathbf{x}_i)$, where $\xi_i$ is a threshold that need to be estimated differently for MCC and MLC. In MCC, for $\forall i \in \{1, \cdots, N\}$, $\xi_i$  should be set to $1$ without estimating. In MLC, a straightforward way to estimate $\xi=[\xi_1, \cdots, \xi_N]$ is to transform the original MLC problem into a general MCC problem to predict the label set size. The details are shown below.

Let $D'= \{\mathbf{x}_i, \xi_i\}_{i=1}^N$ denote the newly-formed training dataset of the transformed MCC problem, and $\mathcal{L}'=unique\left(\{\xi_i\}_{i=1}^N\right)$ forms the transformed label space. Then, the similarity of instances in $\mathcal{D}'$ of the same set size with respect to $\mathbf{x}_i$ is
\begin{equation}
f'_k(\mathbf{x}_i)=\sum_{\mathbf{x}_j \in \mathcal{D}'_k} \Phi \langle \mathbf{x}_i, \mathbf{x}_j\rangle
\end{equation}
where $\mathcal{D}_k' \subseteq \mathcal{D}'$ is the subset of training instances with class $k \in \mathcal{L}'$. Therefore, the set size of $\mathbf{x}_i$ is predicted by the following decision function:
\begin{equation}
\xi(\mathbf{x}_i)={\arg \max}_{k \in \mathcal{Y}'} \sum_{\mathbf{x}_j \in \mathcal{D}'_k} \Phi \langle \mathbf{x}_i, \mathbf{x}_j\rangle
\end{equation}
where $\xi(\mathbf{x}_i)$ is the predicted label set size for $\mathbf{x}_i$. Obviously, $\xi(\mathbf{x}_i)$ is the label set size with maximum similarity.

 The essence of SML is to perform transformation of the feature space. From the perspective of \cite{siblini2019review}, SML is actually a data preprocessing method. Unlike the traditional methods such as PCA \cite{wold1987principal} or LPP \cite{he2004locality}, which maintain the data structural of itself, SML is a data preprocessing method that contains class information. Although simple, it has proven to be effective \cite{rossi2018similarity}. This provides a promising next direction for multi-label data preprocessing. 

\subsection{Similarity-based framework}
As can be seen from the last subsection, SML predicts the class $\lambda_k \in \mathcal{L}$ for unseen instance $\mathbf{x}$ only by using the similarity from instance subset $\mathcal{D}_k \subseteq \mathcal{D}$, ignoring the similarities from other subsets $\{\mathcal{D}_t \subseteq \mathcal{D}\}_{t \in [m]\setminus k}$, where $[m]\setminus k$ denotes the positive integer set $\{1, 2, \cdots, k-1, k+1, \cdots, m\}$.

To compensate such a limitation, our idea is to train one classifier $h_k$ for each class by using all the normalized similarities $f(\mathbf{x}_1), \cdots, f(\mathbf{x}_N)$. Specifically, we use all these similarities as the input to the generalized linear models to predict the class $\lambda_k \in \mathcal{L}$ for unseen instance $\mathbf{x}$. The differences of predicting unseen instance $\mathbf{x}_0$ between our approach and SML are graphically shown in Figure \ref{timec}.
\begin{figure} \centering
\subfigure[SML] {\label{sA1}
\includegraphics[width=0.25\columnwidth]{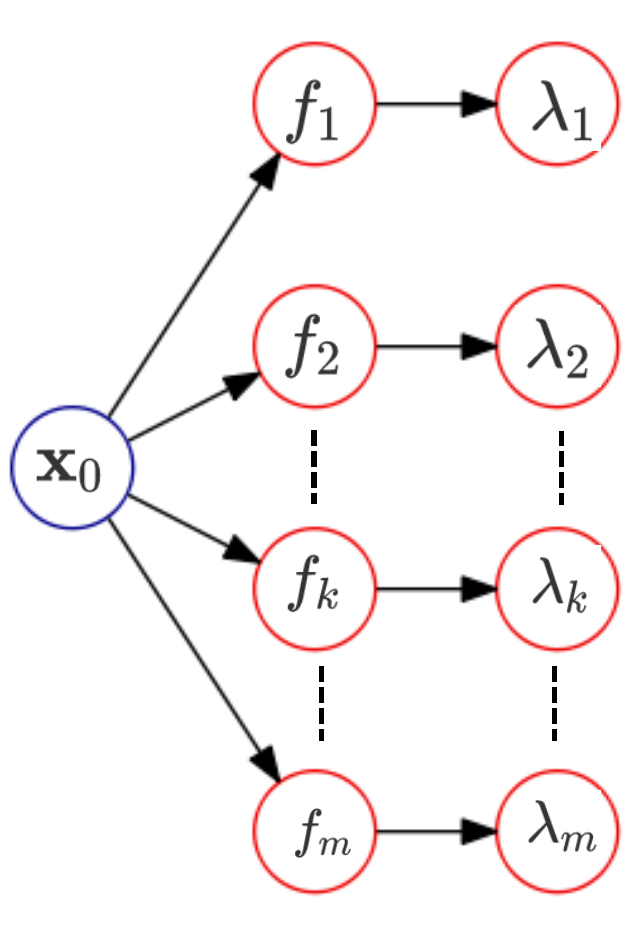}}
\subfigure[Our approach] { \label{sA2}
\includegraphics[width=0.28\columnwidth]{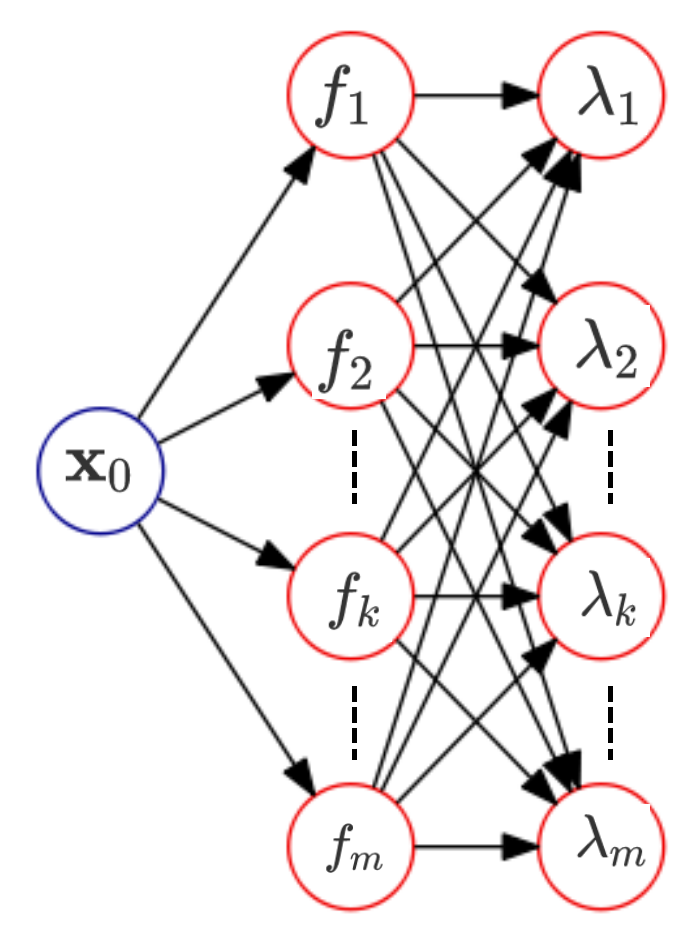}}
\caption{The differences of predicting unseen instance $\mathbf{x}_0$ between SML and our approach. SML predicts the class $\lambda_k$ only by using the similarity evidence $f_k$, while our approach predicts the class $\lambda_k$ by using all similarity evidences $\{f_k\}_{k \in [m]}$}
\label{timec}
\end{figure}

More formally, for the $k$-th class $\lambda_k$, the corresponding classifier is derived from the following generalized linear model
\begin{equation}
\pi^k_0 = g^{-1}\left(\beta_0^k + \sum_{t=1}^m\beta_t^k \bar{f}_t(\mathbf{x_0}) \right)
\end{equation}
where $\pi^k_0$ denotes the expected response for $\mathbf{x}_0$, and $\bar{f}_t(\mathbf{x_0})$ denotes the normalized $f_t(\mathbf{x_0}) = \sum_{\mathbf{x} \in D_t} \Phi \langle \mathbf{x}_0, \mathbf{x} \rangle$, which is the similarity evidence from instance subset $\mathcal{D}_t \subseteq \mathcal{D}$ in favor of $\pi^k(\mathbf{x}_0) = 1$. $g$ is called the link function in the generalized linear model. Ideally, $\pi^k_0 = 1$  if class $\lambda_k$ is relevant for $\mathbf{x}_0$, $0$ otherwise.

Note that various link functions can be chosen depending on the assumed distribution of $\pi^k_0$ \cite{mccullagh1989log}, in this paper, we use the logit link $g(a) = log(a/(1-a))$ for simplicity. This reduces to logistic regression model. Specifically,
\begin{equation}
log\left(\frac{\pi_0^k}{1-\pi_0^k}\right)=\beta_0^k + \sum_{t=1}^m\beta_t^k\bar{f}_t(\mathbf{x_0})
\end{equation}
In this case, $\pi_0^k$ denotes the posterior probability that $\lambda_k$ is relevant for $\mathbf{x}_0$. An optimal specification of $\{\beta_t^k\}_{t=0}^m$ can be accomplished by adapting this parameter to the data $\mathcal{D}$ by using the method of maximum likelihood estimation. The corresponding negative log-likelihood function is then given by
\begin{equation}
\label{classifier_k}
\ell(\boldsymbol{\beta^k})=\sum_{i=1}^n (-y_i^k{\boldsymbol{\beta}^k}^T\mathbf{z}_i+\ln(1+e^{{\boldsymbol{\beta}^k}^T\mathbf{z}_i}))
\end{equation}
where $y_i^k = 1$  if class $\lambda_k$ is relevant for $\mathbf{x}_i$, $0$ otherwise. $\boldsymbol{\beta}^k=[\beta_0^k,\beta_1^k, \cdots, \beta_m^k]^T$ and $\mathbf{z}_i=[1, f_1(\mathbf{x}_i), \cdots, f_m(\mathbf{x}_i)]^T$.

The problem (\ref{classifier_k}) can be computed by means of standard methods from logistic regression \cite{Liu:2009:SLEP:manual}. Here, we refer to this method as SBLR (Similarity-Based multi-label learning by Logistic Regression).

Obviously, these learned parameters $\{\beta_t^k\}_{t \in [m]\setminus k}$ reveal the contribution degree of other classes to class $\lambda_k$. In practice, some similarity evidences from other classes do not provide useful information to class $\lambda_k$, even conversely bring noisy information. To overcome this problem, we impose an $\ell_1$-norm penalty on SBLR to detect the irrelevant classes for preventing from impairing performance of them. This leads to an objective:
\begin{equation}
\label{classifier_sparsek}
\ell(\boldsymbol{\beta^k})=\frac{1}{n}\sum_{i=1}^n (-y^k_i{\boldsymbol{\beta}^k}^T\mathbf{z}_i+\ln(1+e^{{\boldsymbol{\beta}^k}^T\mathbf{z}_i}))
+\lambda\|\boldsymbol{\beta}^k\|_1
\end{equation}
We refer to the model derived from (\ref{classifier_sparsek}) as SparseSBLR (Sparse Similarity-Based multi-label learning by Logistic Regression). It can be solved by means of sparse logistic regression and a useful toolbox is provided by \cite{Liu:2009:SLEP:manual}.

After learning all the parameters $\{\beta_t^k\}_{j=0}^m$ for each class $\lambda_k$, The posterior probabilities $\{\pi_0^k\}_{k \in [m]}$ for the query $\mathbf{x}_0$ are then computed by
\begin{equation}
\pi_0^k=\frac{1}{1+e^{{\boldsymbol{\beta}^k}^T\mathbf{z}_0}},
\end{equation}
then classify $\mathbf{x}_0$ with applying the decision rule
\begin{equation}
h(\mathbf{x_0})=\left\{\lambda_k \in \mathcal{L} | \pi_0^k \geq 1/2, k \in [m] \right\}
\end{equation}
for MLC, and the decision rule
\begin{equation}
h(\mathbf{x_0})=\arg \max_{\lambda_k} \{\pi_0^k\}_{k \in [m]}
\end{equation}
for MCC, respectively.
\subsection{Complexity analysis}
In this subsection, we detail the computational complexities of the proposed methods.
Given $N$ training instances and a test instance $\mathbf{x}_0$, the time complexity of constructing the similarity evidences for all training instances is $\mathcal{O}(dN^2)$. Then, the time complexity for training SBLR is $\mathcal{O}(\eta Nm^2)$ if we use the gradient descent method, where $\eta$ is the number of iterations. And training SparseSBLR costs also $\mathcal{O}(\eta Nm^2)$ if using the accelerated proximal gradient method. The time complexity for computing posterior probabilities $\{\pi_0^k\}_{k \in [m]}$ for $\mathbf{x}_0$ is $\mathcal{O}(m^2)$.		
In a nutshell, the total time complexities for training SBLR and SparseSBLR are both $\mathcal{O}(\eta Nm^2+dN^2)$. The time complexities of SBLR and SparseSBLR for predicting $\mathbf{x}_0$ are both $\mathcal{O}(dN+m^2)$.

\section{Experiments}
Due to the effectiveness of using the sum of similarities from the instances of each class as features has been validated in \cite{rossi2018similarity}, in this section, we focus on demonstrating the effectiveness of the proposed methods, i.e., SparseSBLR and SBLR, on both MLC and MCC datasets. The codes of our approaches are publicly availabel on \url{https://github.com/John986/paperCodes}.

\subsection{Experiments on MLC datasets}
We use five commonly MLC criteria, including Hamming loss, one error, Coverage, Rank loss and Average precision \cite{zhang2014review}. For each criterion, we use $\uparrow$($\downarrow$) to denote the larger (smaller) the value, the better the performance. Before presenting the results of our experiments, we detail the datasets, compared methods used for evaluation.  
\subsubsection{Datasets}
We use seven real-world datasets \footnote{http://mulan.sourceforge.net/datasets-mlc.html} and give an overview of these datasets in Table \ref{s1}.

\begin{table}[]
\scriptsize
\centering
\caption{Statistics for the MLC datasets used in the experiments. The symbol \#A indicates the numbers of A; cardinality is the average number of labels per instance.}
\label{s1}
\begin{tabular}{llllll}
\hline
Data set & Domain  & \#Instances & \#Attributes & \#Labels & Cardinality \\ \hline \hline
emotions & music   & 593         & 72           & 6        & 1.869       \\
scene    & image   & 2407        & 294          & 6        & 1.074       \\
yeast    & biology & 2417        & 103          & 14       & 4.237       \\
birds    & audio   & 645         & 260          & 19       & 1.014       \\
genbase  & biology & 662         & 1186         & 27       & 1.252       \\
medical  & text    & 978         & 1449         & 45       & 1.245       \\
CAL500   & music   & 502         & 68           & 174      & 26.044      \\\hline
\end{tabular}
\end{table}

\subsubsection{Compared methods}
We compare the proposed methods SBLR and SparseSBLR to three MLC methods:

1) BR-SVM \cite{luaces2012binary} works by training a number of independent SVM classifiers, one per class space. Therefore, BR-SVM does not consider dependencies among class spaces in model induction.

2) SML \cite{rossi2018similarity}, as reviewed in the introduction, gives rise to a new class of methods for MLC and also achieves promising performance. The code of SML is not publicly available. Thus, we implement this algorithm by ourselves and the code is publicly availabel on \url{https://github.com/John986/paperCodes}}\footnote{There are some differences between the results of our SML implemention and the original results reported in \cite{rossi2018similarity} on the yeast dataset, which may be brought by different data preprocessing methods.}.

3) MLKNN \cite{zhang2007ml} learns a single classifier for each label by means of a combination of KNN and Bayesian inference, and still can be considered as the state-of-the-art MLC classifier. It is parameterized by the size $k$ of the neighborhood, for which we adopted $k=10$ as recommended in \cite{zhang2007ml}, the value yields the best performance. This method is implemented by using the codes directly provided by the authors.

4) IBLR\footnote{In \cite{cheng2009combining}, IBLR can beat logistic regression model on MLC datasets. Thus, logistic regression model is not used as compared method here.} \cite{cheng2009combining} learns a single classifier for each label by combining KNN and logistic regression. It is also parameterized by the size of the neighborhood, for which we adopted the same value as stated in \cite{cheng2009combining}. Likewise, we use its implementation in the MULAN package \cite{Tsoumakas2010MiningMD}.

BR-SVM, SML and our proposed methods SBLR, SparseSBLR all use the RBF similarity function. The RBF hyperparameter and the hyperparameter $\lambda$ of SparseSBLR are both selected from the set $\{10^2, 10^1, 10^0, 10^{-1}, 10^{-2}, 10^{-3}, 10^{-4}, 10^{-5}\}$ via cross-validation on $10\%$ of the training data.

\subsubsection{Results}
\begin{table}[]
\tiny
\centering
\caption{Experimental results in terms of Hamming Loss.}
\label{Hamming-Loss}
\begin{tabular}{lllllll}
\hline
Hamming Loss $\downarrow$ & SparseSBLR        & SBLR     & SML      & MLKNN             & IBLR              & BR-SVM   \\ \hline
emotions                  & 0.190(2)          & 0.196(4) & 0.238(6) & 0.195(3)          & \textbf{0.186(1)} & 0.198(5) \\
scene                     & 0.112(4)          & 0.094(2) & 0.119(6) & \textbf{0.087(1)} & 0.113(5)          & 0.111(3) \\
yeast                     & \textbf{0.193(1)} & 0.196(4) & 0.218(6) & 0.195(3)          & 0.193(2)          & 0.199(5) \\
birds                     & \textbf{0.042(1)} & 0.042(2) & 0.053(5) & 0.047(3)          & 0.048(4)          & 0.053(6) \\
genbase                   & \textbf{0.001(1)} & 0.002(3) & 0.009(6) & 0.005(5)          & 0.003(4)          & 0.001(2) \\
medical                   & 0.042(4)          & 0.030(3) & 0.094(6) & 0.05(5)           & \textbf{0.021(1)} & 0.028(2) \\
CAL500                    & \textbf{0.135(1)} & 0.141(5) & 0.137(2) & 0.139(3)          & 0.172(6)          & 0.139(4) \\ \hline
Average rank              & \textbf{2.00}     & 3.29     & 5.29     & 3.29              & 3.29              & 3.86     \\ \hline
\end{tabular}
\end{table}

\begin{table}[]
\tiny
\centering
\caption{Experimental results in terms of Ranking Loss.}
\label{Ranking-Loss}
\begin{tabular}{lllllll}
\hline
Rank Loss $\downarrow$ & SparseSBLR & SBLR  & SML   & MLKNN & IBLR & BR-SVM\\ \hline
emotions                  & \textbf{0.147(1)} & 0.148(2)          & 0.156(6) & 0.154(5) & 0.151(3)          & 0.153(4) \\
scene                     & 0.079(2)          & \textbf{0.078(1)} & 0.097(5) & 0.079(3) & 0.116(6)          & 0.094(4) \\
yeast                     & 0.166(2)          & 0.17(4)           & 0.178(5) & 0.168(3) & \textbf{0.164(1)} & 0.199(6) \\
birds                     & 0.15(2)           & 0.179(4)          & 0.216(6) & 0.158(3) & \textbf{0.084(1)} & 0.182(5) \\
genbase                   & \textbf{0.001(1)} & 0.003(3)          & 0.015(6) & 0.007(5) & 0.006(4)          & 0.001(2) \\
medical                   & \textbf{0.02(1)}  & 0.023(2)          & 0.087(6) & 0.037(4) & 0.038(5)          & 0.026(3) \\
CAL500                    & \textbf{0.179(1)} & 0.184(4)          & 0.18(2)  & 0.183(3) & 0.231(5)          & 0.27(6)  \\ \hline
Average rank              & \textbf{1.43}     & 2.86              & 5.14     & 3.71     & 3.57              & 4.29     \\ \hline
\end{tabular}
\end{table}

\begin{table}[]
\tiny
\centering
\caption{Experimental results in terms of One Error.}
\label{One-Error}
\begin{tabular}{lllllll}
\hline
One Error   $\downarrow$  & SparseSBLR & SBLR  & SML   & MLKNN & IBLR & BR-SVM \\ \hline
emotions                  & \textbf{0.237(1)} & 0.268(6)          & 0.241(2) & 0.264(5)          & 0.256(4) & 0.249(3) \\
scene                     & 0.236(2)          & 0.238(3)          & 0.269(5) & \textbf{0.225(1)} & 0.319(6) & 0.258(4) \\
yeast                     & \textbf{0.228(1)} & 0.233(5)          & 0.249(6) & 0.228(2)          & 0.230(4) & 0.228(3) \\
birds                     & 0.388(2)          & \textbf{0.383(1)} & 0.571(5) & 0.465(4)          & 0.697(6) & 0.403(3) \\
genbase                   & \textbf{0.002(1)} & 0.002(2)          & 0.027(6) & 0.015(4)          & 0.015(5) & 0.006(3) \\
medical                   & 0.087(2)          & \textbf{0.081(1)} & 0.355(6) & 0.165(4)          & 0.235(5) & 0.087(3) \\
CAL500                    & \textbf{0.116(1)} & 0.146(4)          & 0.118(3) & 0.116(2)          & 0.422(6) & 0.314(5) \\ \hline
Average rank              &\textbf{1.43}             & 3.14              & 4.71     & 3.14              & 5.14     & 3.43     \\ \hline
\end{tabular}
\end{table}

\begin{table}[]
\tiny
\centering
\caption{Experimental results in terms of Coverage.}
\label{Coverage}
\begin{tabular}{lllllll}
\hline
Coverage   $\downarrow$  & SparseSBLR & SBLR    & SML     & MLKNN   & IBLR & BR-SVM \\ \hline
emotions                  & 1.710(2)            & 1.717(3)          & 1.759(6)   & 1.742(5)          & \textbf{1.707(1)} & 1.737(4)   \\
scene                     & 0.480(2)            & \textbf{0.475(1)} & 0.569(5)   & 0.481(3)          & 0.665(6)          & 0.558(4)   \\
yeast                     & \textbf{6.171(1)}   & 6.241(3)          & 6.415(5)   & 6.298(4)          & 6.195(2)          & 7.169(6)   \\
birds                     & 2.106(2)            & 2.423(3)          & 2.761(6)   & \textbf{1.981(1)} & 2.434(4)          & 2.556(5)   \\
genbase                   & \textbf{0.266(1)}   & 0.385(3)          & 0.889(6)   & 0.562(5)          & 0.489(4)          & 0.288(2)   \\
medical                   & \textbf{0.409(1)}   & 0.427(2)          & 1.056(5)   & 0.563(4)          & 3.464(6)          & 0.473(3)   \\
CAL500                    & \textbf{129.026(1)} & 131.278(4)        & 129.476(2) & 131.04(3)         & 133.644(5)        & 156.014(6) \\ \hline
Average rank              & \textbf{1.43}       & 2.71              & 5.00       & 3.57              & 4.00              & 4.29       \\ \hline
\end{tabular}
\end{table}

In the experiments, we adopt a 10-fold cross-validation strategy to compute the mean of classification performance. The results are presented in Tables \ref{Hamming-Loss}-\ref{Average-Precision}, where the number in brackets behind the performance value is the rank of the method on the corresponding data set and the average rank is the average of the ranks across all datasets.

\begin{table}[]
\tiny
\centering
\caption{Experimental results in terms of Average Precision.}
\label{Average-Precision}
\begin{tabular}{lllllll}
\hline
Average Precision $\uparrow$ & SparseSBLR & SBLR  & SML   & MLKNN & IBLR & BR-SVM \\ \hline
emotions                  & \textbf{0.820(1)} & 0.808(5) & 0.81(4)  & 0.807(6)          & 0.812(3)          & 0.813(2) \\
scene                     & 0.860(2)          & 0.860(3) & 0.837(5) & \textbf{0.865(1)} & 0.809(6)          & 0.843(4) \\
yeast                     & 0.766(2)          & 0.764(3) & 0.742(6) & 0.764(4)          & \textbf{0.769(1)} & 0.749(5) \\
birds                     & \textbf{0.654(1)} & 0.635(2) & 0.509(6) & 0.612(5)          & 0.615(4)          & 0.618(3) \\
genbase                   & \textbf{0.998(1)} & 0.996(2) & 0.970(6) & 0.985(4)          & 0.983(5)          & 0.995(3) \\
medical                   & \textbf{0.945(1)} & 0.928(3) & 0.776(6) & 0.900(4)          & 0.809(5)          & 0.944(2) \\
CAL500                    & \textbf{0.504(1)} & 0.502(2) & 0.501(3) & 0.492(4)          & 0.396(6)          & 0.445(5) \\ \hline
Average rank              & \textbf{1.29}     & 2.86     & 5.14     & 4.00              & 4.29              & 3.43     \\ \hline
\end{tabular}
\end{table}

As we can see from the above tables, our proposed model SparseSBLR, achieves the best average ranking in terms of Hamming loss, Rank loss,  One error, Coverage and Average precision. More specifically, 1) both SparseSBLR and SBLR achieve better average rank performance than SML in terms of all evaluation measures, which verifies that incorporating class dependencies into SML is useful to further promote its classification performance for MLC; 2) compared to the compared methods, SparseSBLR nearly achieves best performance on the datasets with relatively more classes, i.e., birds, genbase, medical and CAL500. This phenomenon indicates that SparseSBLR can indeed detect the irrelevant classes and prevent them from impairing classification performance.

\begin{figure}
\setlength{\abovecaptionskip}{0cm}
\setlength{\belowcaptionskip}{0cm}
\centering
\subfigure[Hamming Loss] {\label{hamming}
\includegraphics[width=0.48\columnwidth]{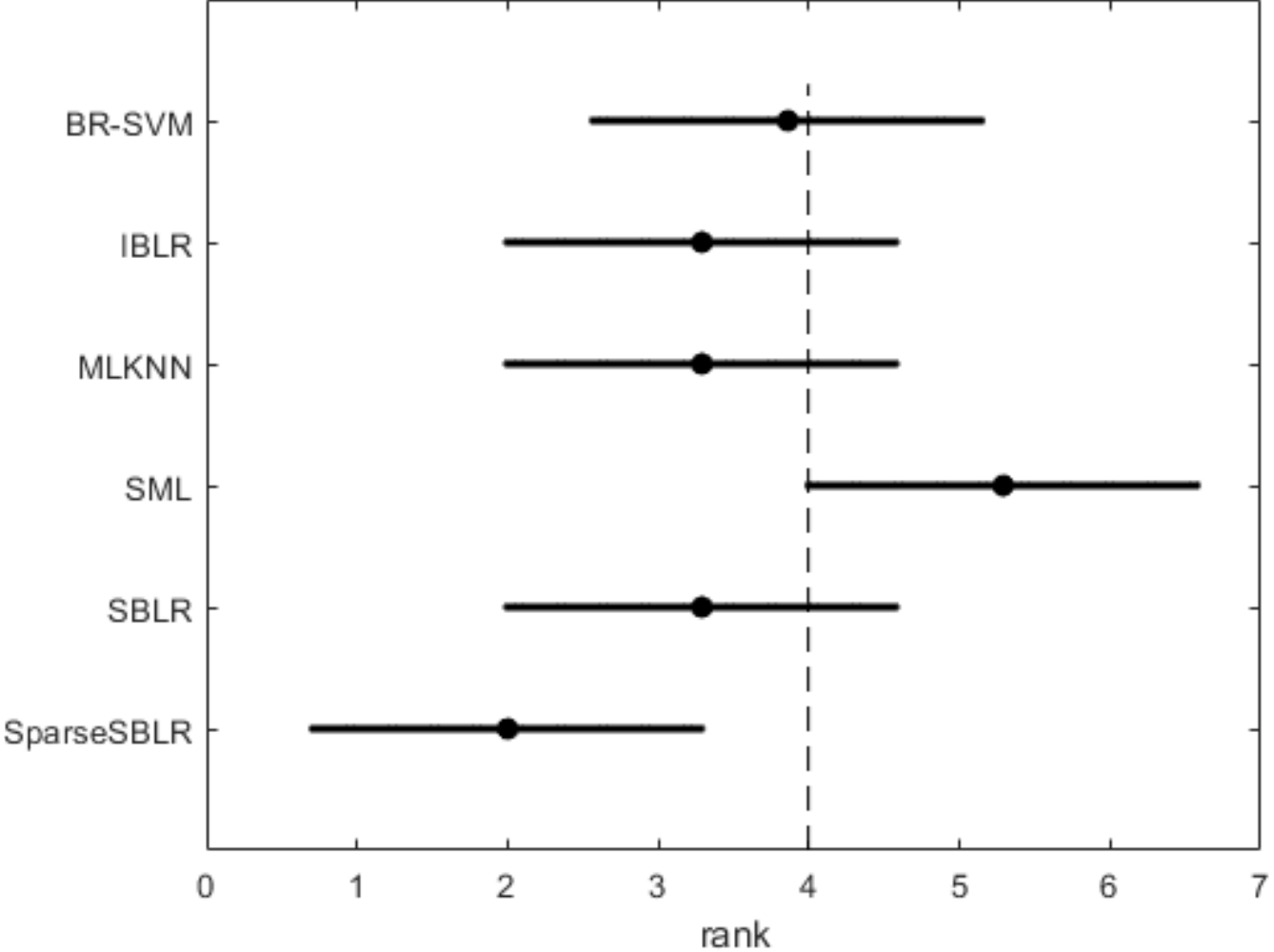}}
\subfigure[Rank Loss] { \label{ranking}
\includegraphics[width=0.48\columnwidth]{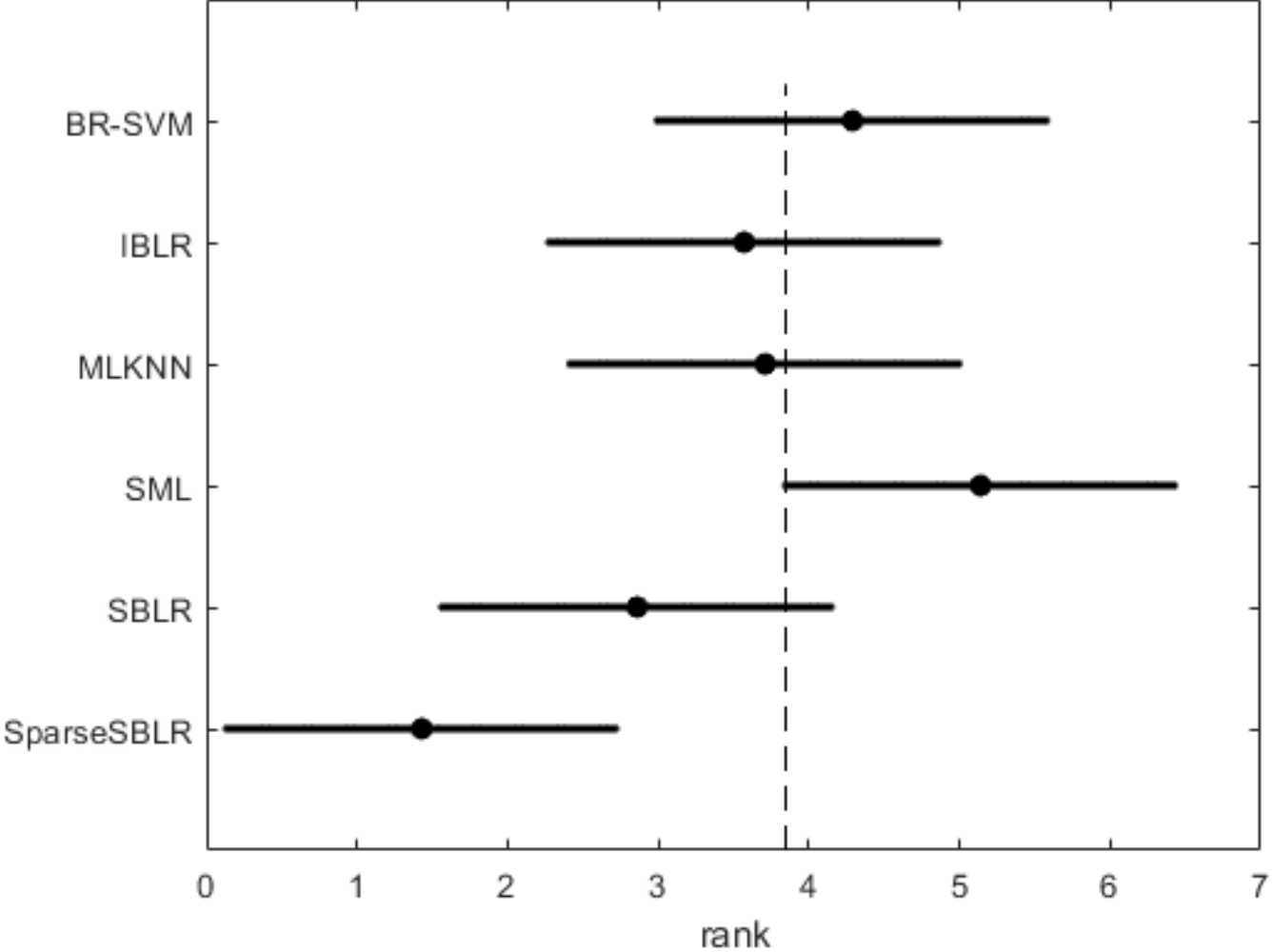}}
\subfigure[One Error Loss] { \label{oneerror}
\includegraphics[width=0.48\columnwidth]{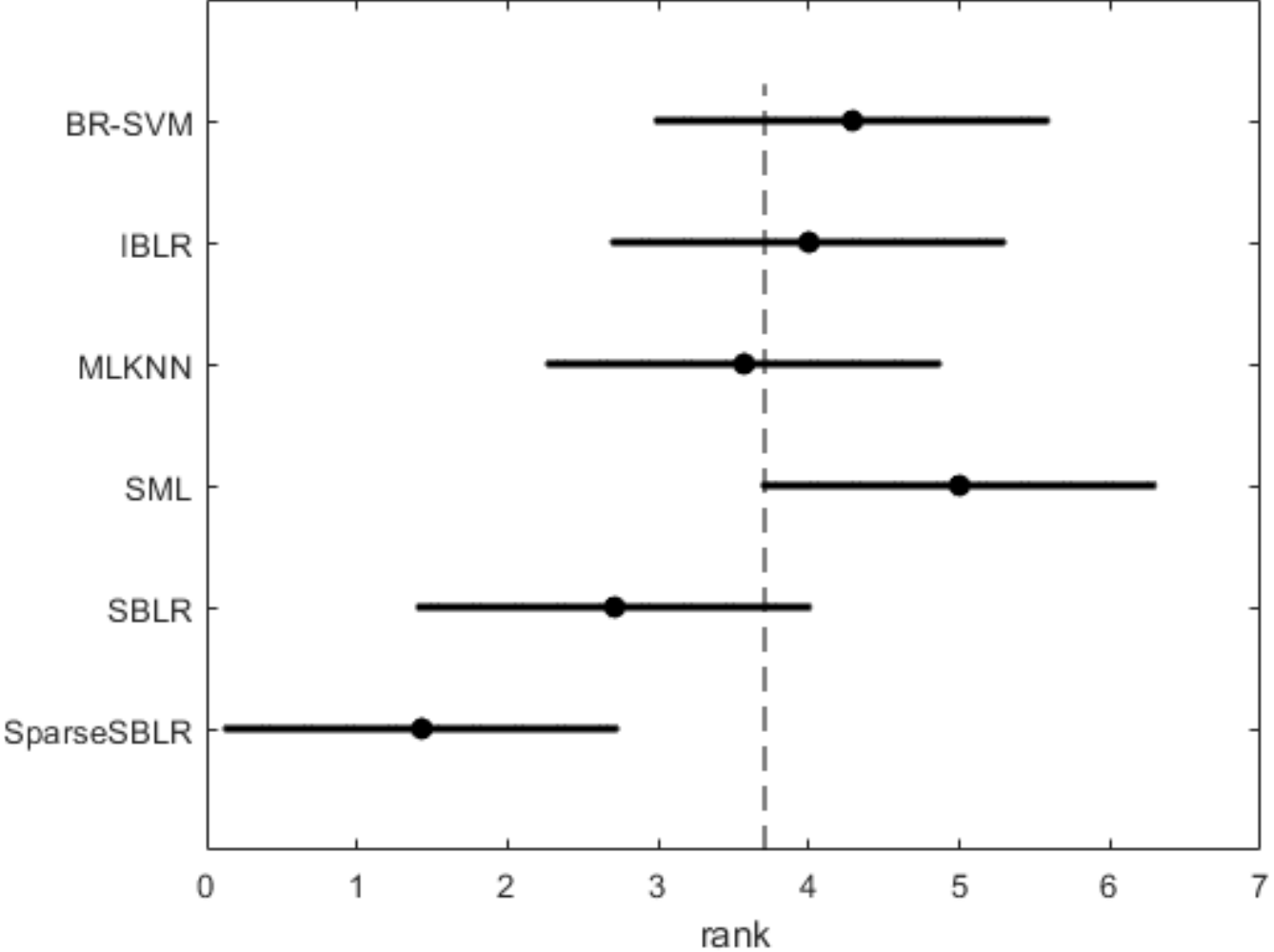}}
\subfigure[Coverage Loss] { \label{converage}
\includegraphics[width=0.48\columnwidth]{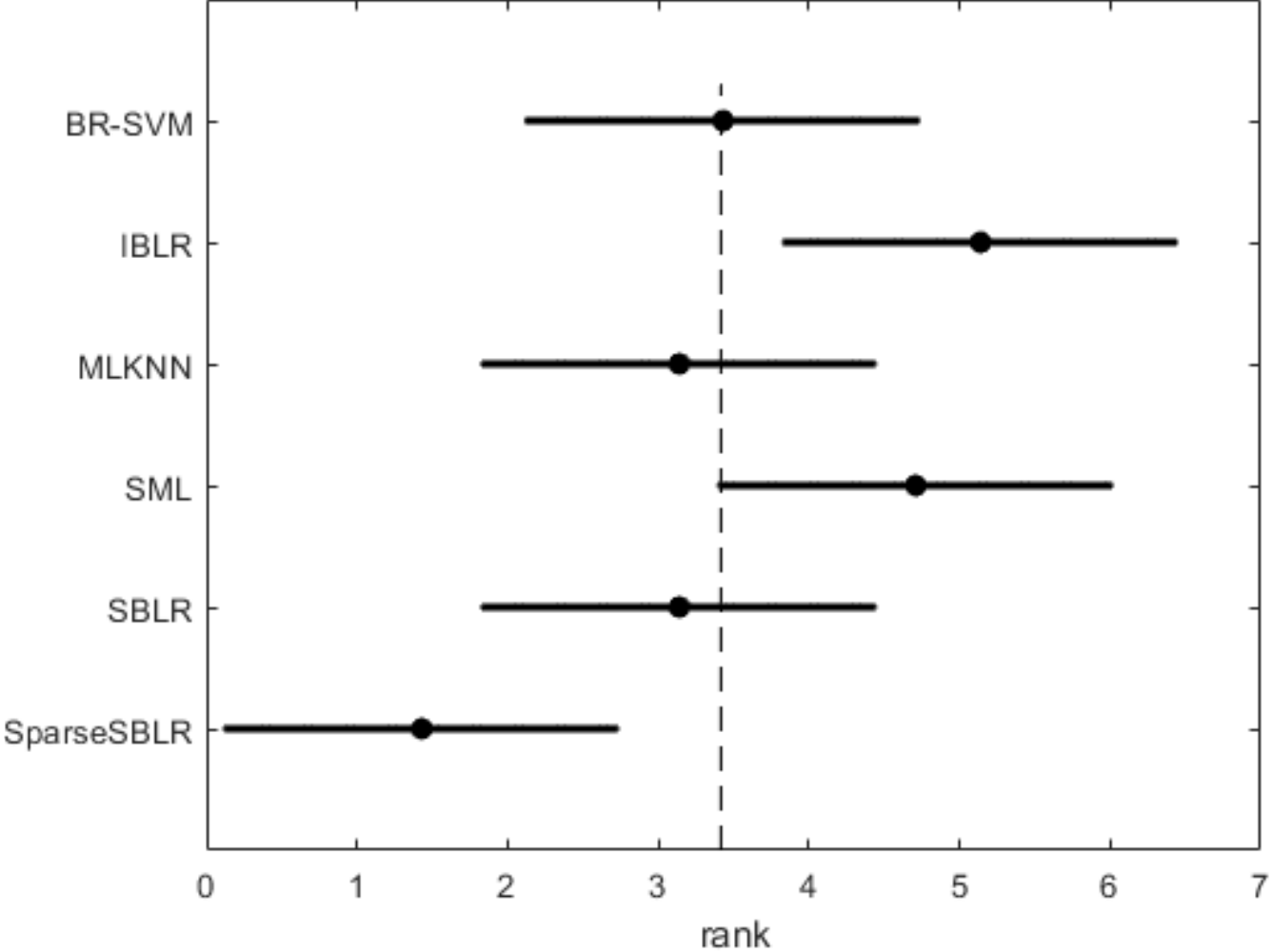}}
\subfigure[Average Precision] { \label{averageprecision}
\includegraphics[width=0.48\columnwidth]{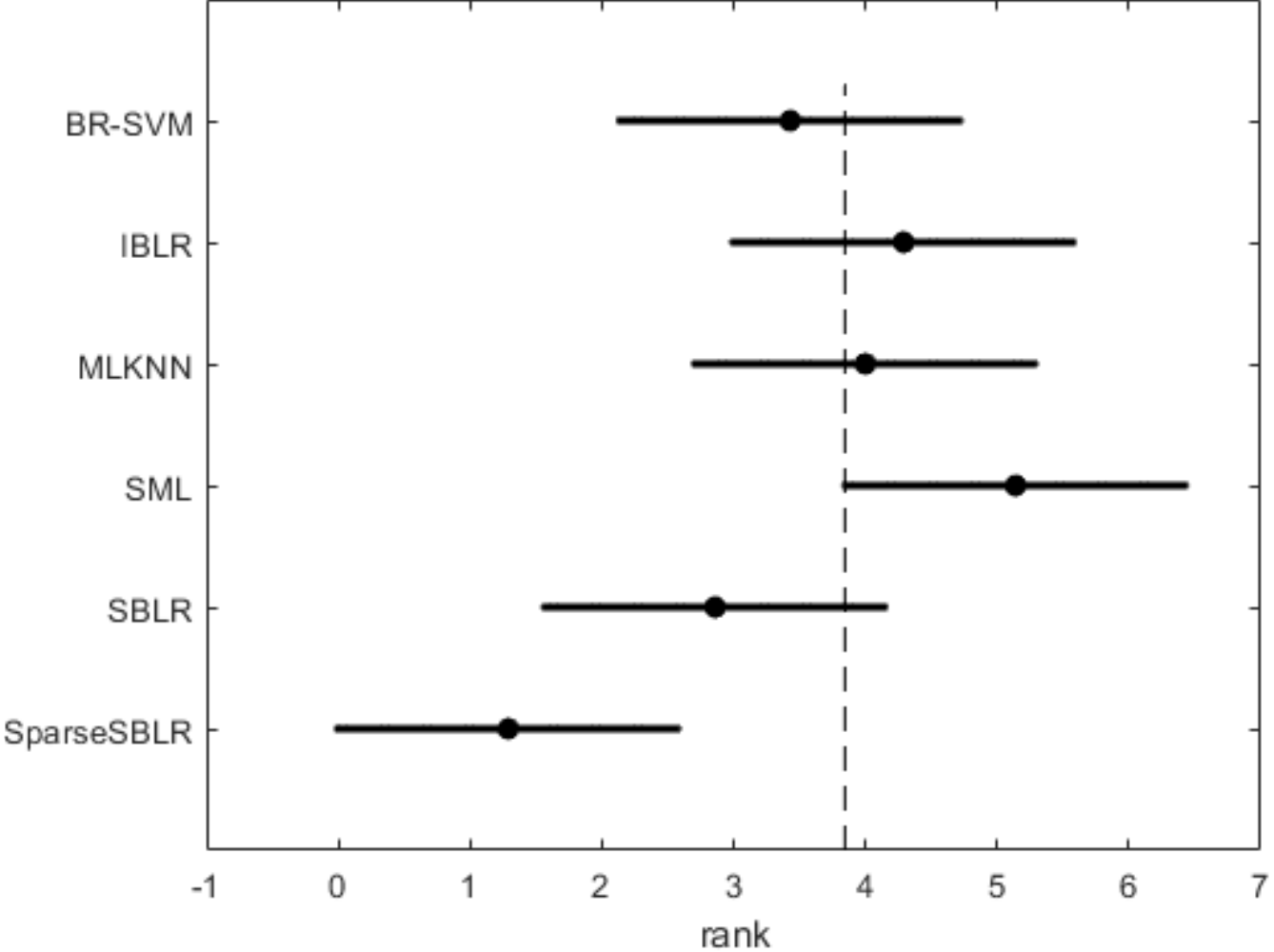}}
\subfigure[Classification Accuracy] { \label{accuracy}
\includegraphics[width=0.48\columnwidth]{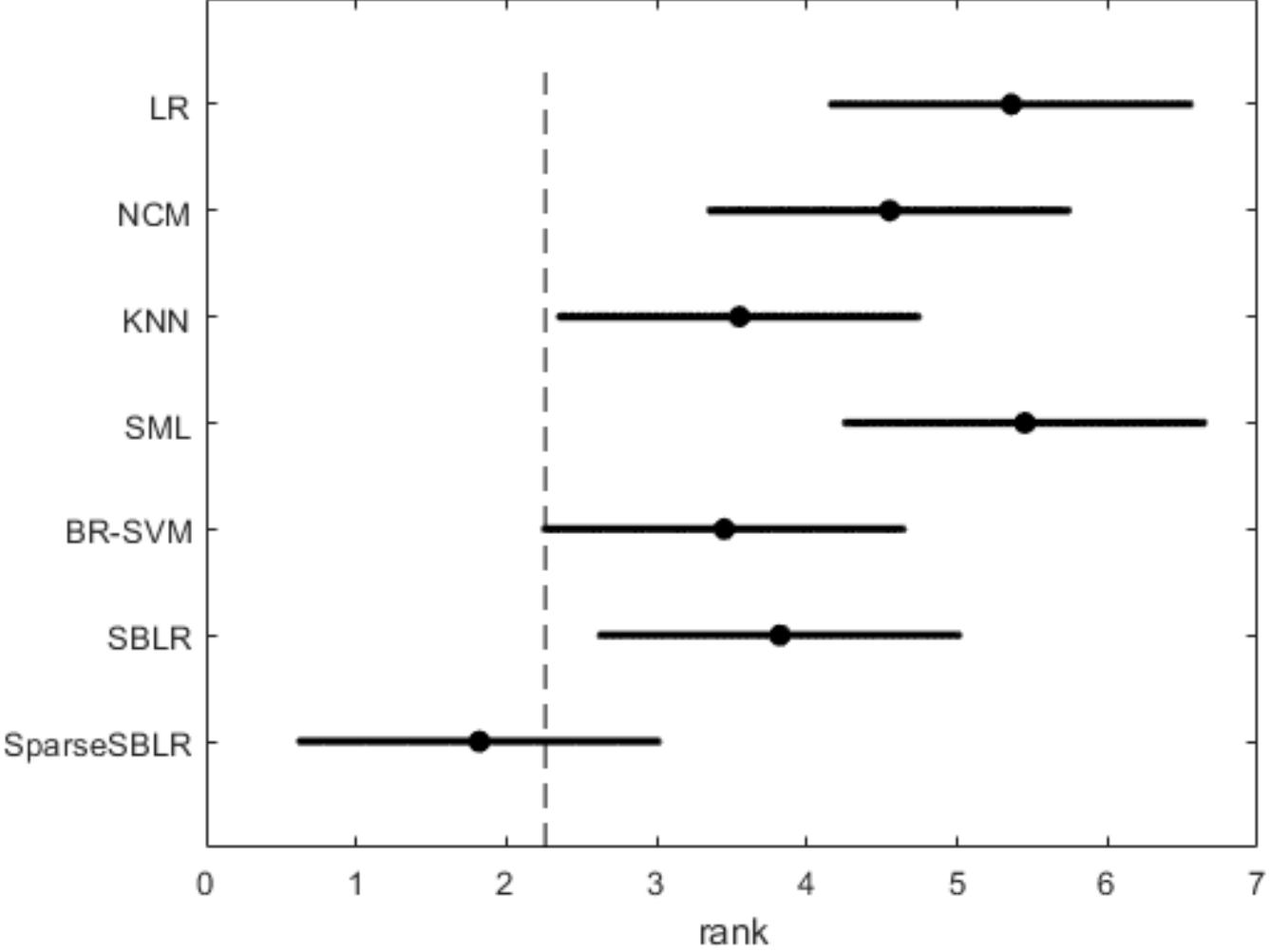}}
\caption{In the graph, the horizontal axis represents the values of mean rank, the vertical axis represents the different methods for test. For each method, the $\bullet$ represents its mean rank value, the line segment represents the critical range of Nemenyi test. Two methods have a significant difference, if their line segments are not overlapped; not so, otherwise.}
\label{nemenyitest}
\end{figure}
To verify whether the differences between different methods are significant in terms of all evaluation measures, the Nemenyi tests are conducted. The results are shown in Figures \ref{hamming}-\ref{averageprecision} from which we can see that, in Hamming loss, Rank loss, One-error, Coverage and Average precision, SparseSBLR is significantly different from SML, while SBLR, BR-SVM, MLKNN and IBLR achieve comparable performance with SML. Thus, we can conclude that SparseSBLR achieves excellent performance on all evaluation measures across all the datasets here.
 

\subsection{Experiments on MCC datasets}
Similar to the above subsection, we also give the experimental datasets and compared methods to evaluate the efficacy of the proposed methods on MCC task.
\subsubsection{Datasets}
We use eleven datasets from UCI machine learning repository \cite{frank2010uci} and give an overview of them in Table \ref{statistic_d2}.

\begin{table}[]
\tiny
\centering
\caption{Statistics for the MCC datasets used in the experiments. The symbol \#A indicates the numbers of A.}
\label{statistic_d2}
\begin{tabular}{llll}
\hline
dataset     & \#classes & \#instances & \#features \\ \hline \hline
Balance     & 3         & 625         & 4          \\
Cmc         & 3         & 1473        & 9          \\
Tae         & 3         & 151         & 5          \\
Wine        & 3         & 178         & 13         \\
Thyroid     & 3         & 215         & 5          \\
Vehicle     & 4         & 846         & 18         \\
Dermatology & 6         & 366         & 33         \\
Glass       & 6         & 214         & 10         \\
Zoo         & 7         & 101         & 17         \\
Ecoli       & 8         & 336         & 8          \\
Vowel       & 11        & 990         & 10         \\ \hline
\end{tabular}
\end{table}

\begin{table}[]
\tiny
\centering
\caption{Experimental results in terms of classification accuracy.}
\label{mcc_re}
\begin{tabular}{llllllll}
\hline
Accuracy    $\uparrow$ & SparseSBLR        & SBLR              & BR-SVM   & SML      & kNN               & NCM      & LR       \\ \hline
balance                & 0.913(2)          & \textbf{0.913(1)} & 0.882(3) & 0.877(4) & 0.873(5)          & 0.742(7) & 0.857(6) \\
cmc                    & \textbf{0.486(1)} & 0.482(2)          & 0.480(3) & 0.388(5) & 0.368(6)          & 0.463(4) & 0.365(7) \\
tae                    & \textbf{0.633(1)} & 0.593(4)          & 0.493(6) & 0.600(3) & 0.62(2)           & 0.553(5) & 0.487(7) \\
wine                   & \textbf{0.994(1)} & 0.853(7)          & 0.965(3) & 0.941(6) & 0.953(4)          & 0.953(5) & 0.965(2) \\
thyroid                & \textbf{0.976(1)} & 0.948(6)          & 0.962(3) & 0.948(5) & 0.962(2)          & 0.952(4) & 0.800(7) \\
vehicle                & 0.663(6)          & 0.664(5)          & 0.779(2) & 0.554(7) & 0.710(3)          & 0.710(4) & \textbf{0.791(1)} \\
dermatology            & \textbf{0.977(1)} & 0.971(4)          & 0.974(2) & 0.914(6) & 0.971(3)          & 0.960(5) & 0.854(7) \\
glass                  & \textbf{0.924(1)} & 0.867(2)          & 0.843(3) & 0.567(7) & 0.743(4)          & 0.743(5) & 0.681(6) \\
zoo                    & 0.960(4)          & 0.950(6)          & 0.960(3) & 0.930(7) & \textbf{0.970(1)} & 0.970(2) & 0.950(5) \\
ecoli                  & \textbf{0.885(1)} & 0.858(2)          & 0.809(3) & 0.770(6) & 0.764(7)          & 0.788(4) & 0.773(5) \\
vowel                  & \textbf{0.981(1)} & 0.828(3)          & 0.462(7) & 0.698(4) & 0.974(2)          & 0.591(5) & 0.584(6) \\ \hline
Average rank           & \textbf{1.82}            & 3.82            & 3.45   & 5.45   & 3.55          & 4.55   & 5.36   \\ \hline
\end{tabular}
\end{table}
\subsubsection{Compared methods}
We compare the proposed methods SBLR and SparseSBLR to four MCC methods:

1) BR-SVM \cite{luaces2012binary} works by training a number of independent SVM classifiers, one per class. The class with the largest score is the final predicted output.

2) Logistic Regression model (LR) \cite{anzai2012pattern} is an inductive classification method and used as a baseline for MCC task.

3) KNN \cite{altman1992introduction} is a simple but effective method for classification. An instance is classified by a majority vote of its neighbors, with the instance being assigned to the class most common among its $k$ nearest neighbors.

4) NCM \cite{schutze2008introduction} is a classification model that assigns to an instance the class of training instances whose mean is closest to the instance.

5) SML \cite{rossi2018similarity} has been used as a compared method for MLC task.  Because it can be naturally applied to MCC, we also used here as a compared method.

BR-SVM, SML, SBLR and SparseSBLR still use the RBF similarity function. And, their hyperparameters are also chosen in the same way as that in the MLC experiments.

\subsubsection{Results}
In the experiments, we adopt a 10-fold cross-validation strategy to compute the mean of classification accuracy. The classification results are presented in Table \ref{mcc_re}. From the Table \ref{mcc_re}, we can see that SparseSBLR and SBLR achieve the highest and the third-highest average ranking, respectively. To verify whether the differences between different methods are significant, the Nemenyi tests are conducted. The results are shown in Figure \ref{accuracy}, from which we can see that SparseSBLR is significantly different from SML, while SBLR, KNN, NCM and LR achieve comparable performance with SML. Therefore, we can conclude that SparseSBLR achieves excellent classification performance in MCC.

\subsection{Discussion}
\begin{table}[]
\tiny
\centering
\caption{The learned parameters of generalized linear models with respect to class 'Pacific Wren' on birds dataset. \#co-occurrence represent the counts of training instances being both relevant to class 'Pacific Wren' and the class in each row.}
\label{birds_d}
\begin{tabular}{cccc}
\hline
Class                     & \#co-occurrence & \multicolumn{1}{c}{SBLR} & \multicolumn{1}{c}{SparseSBLR} \\ \hline
Pacific Wren              & 74              & 20.59                    & 7.92                           \\
Swainson's Thrush         & 18              & 0.10                     & -3.23                          \\
Hammond Flycatcher        & 13              & -0.90                    & -1.07                          \\
Pacific-slope Flycatcher  & 11              & -1.29                    & -6.29                          \\
Western Tanager           & 11              & -0.38                    & 0.05                           \\
Chestnut-backed Chickadee & 5               & 0.54                     & 0.00                           \\
Varied Thrush             & 5               & -1.25                    & 2.25                           \\
Hermit Warbler            & 5               & -0.87                    & 0.00                           \\
Golden Crowned Kinglet    & 5               & -0.34                    & 0.92                           \\
Olive-sided Flycatcher    & 2               & -0.26                    & 1.11                           \\
Hermit Thrush             & 1               & 0.17                     & -0.80                          \\
Stellar's Jay             & 1               & -0.05                    & -0.57                          \\
Common Nighthawk          & 1               & 0.04                     & 0.00                           \\
Brown Creeper             & 0               & 0.12                     & 0.00                           \\
Red-breasted Nuthatch     & 0               & 0.45                     & 0.00                           \\
Dark-eyed Junco           & 0               & -1.45                    & 0.00                           \\
Black-headed Grosbeak     & 0               & 0.40                     & 0.07                           \\
Warbling Vireo            & 0               & 0.66                     & 0.00                           \\
MacGillivray's Warbler    & 0               & 0.03                     & -0.27                          \\ \hline
\end{tabular}
\end{table}

\begin{figure}
	\centering
	\includegraphics[width=0.35\textwidth]{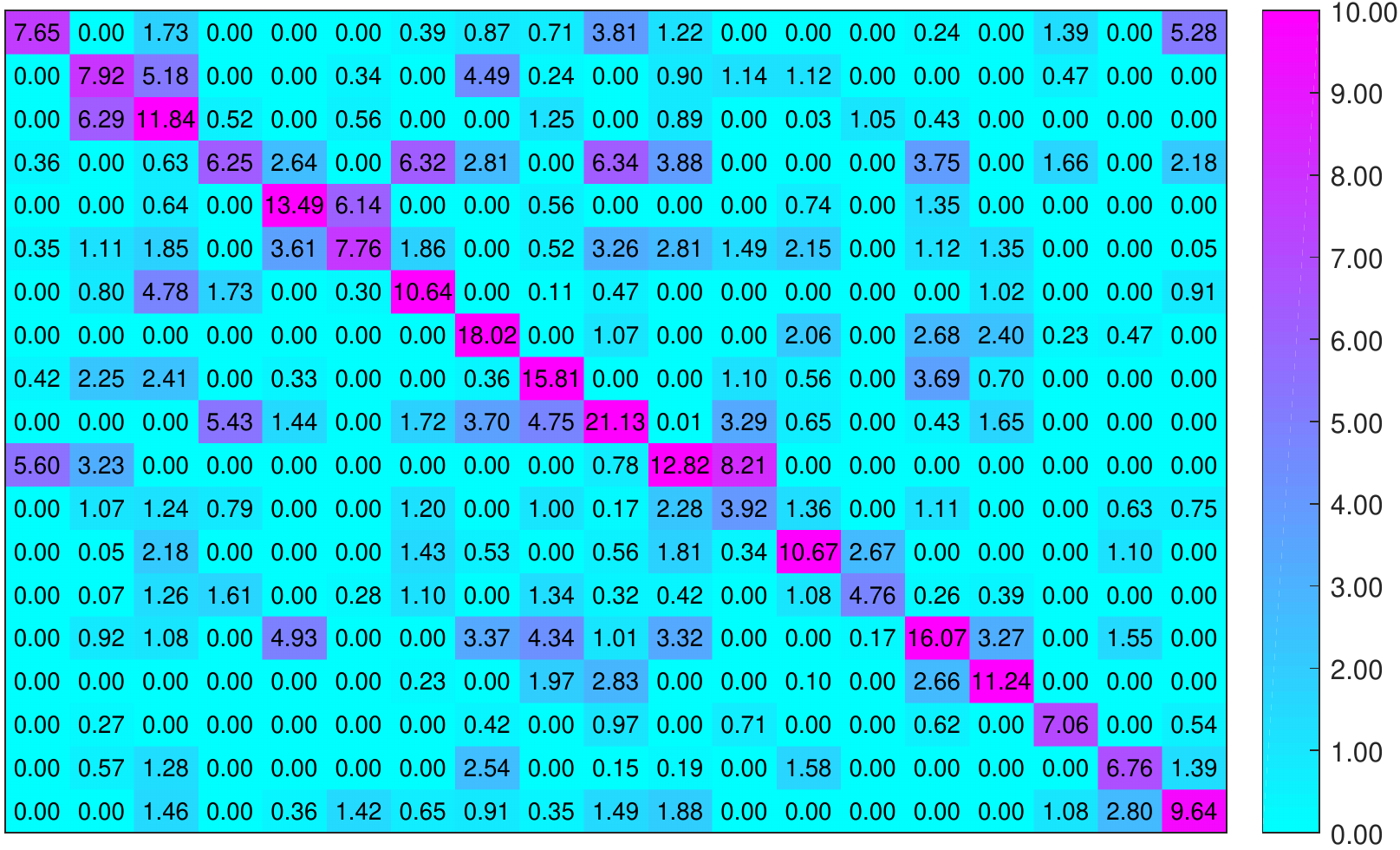}
	\caption{The absolute values of the SparseSBLR parameter on the birds dataset. }
	\label{heatmap_sparsesblr}       
\end{figure}

\begin{figure}
	\centering
	\includegraphics[width=0.33\textwidth]{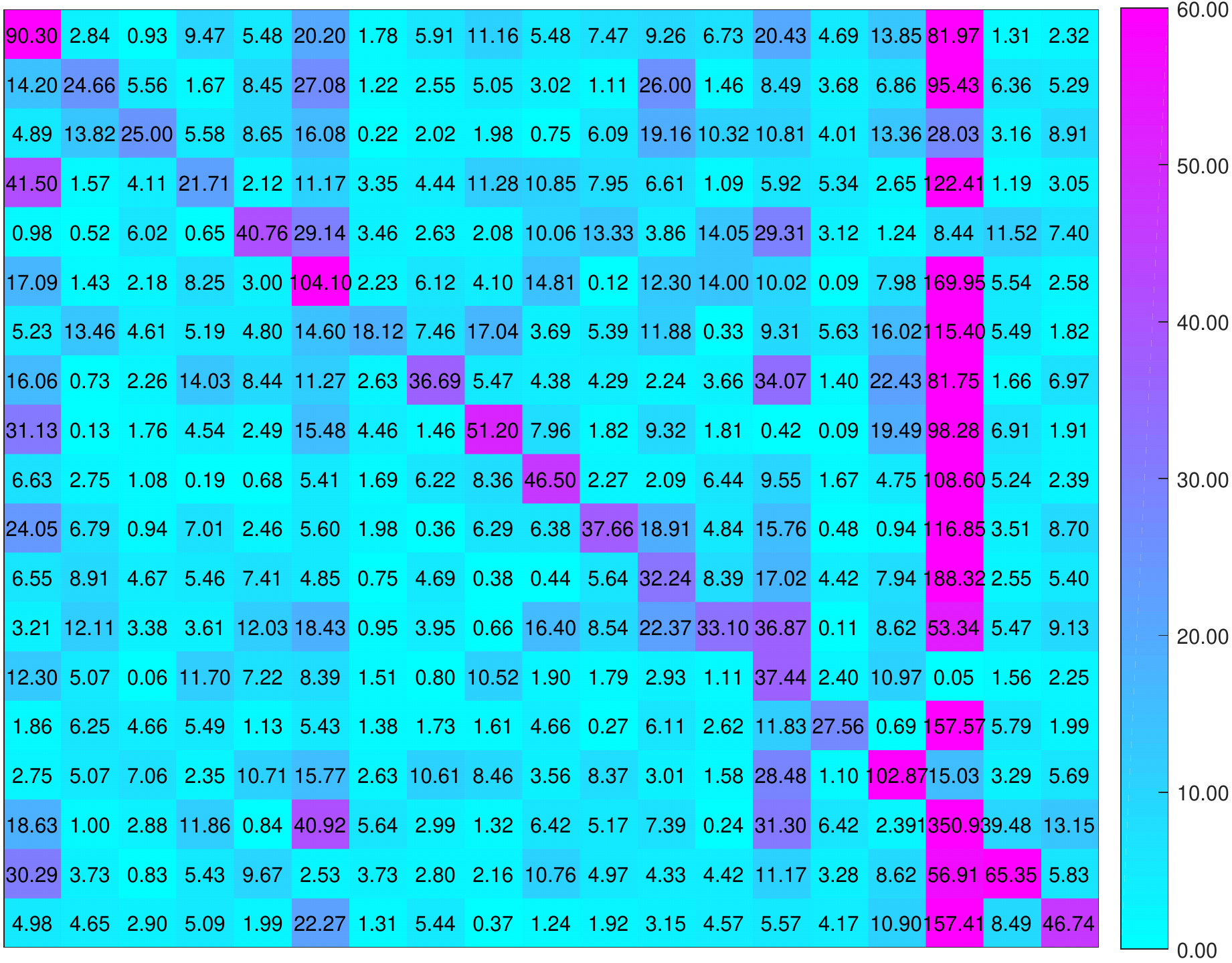}
	\caption{The absolute values of the SBLR parameter on the birds dataset.}
	\label{heatmap_sblr}       
\end{figure}

The learned parameters of generalized linear models indicate the contribution of one class to another class, providing interpretability to some extent. We show the absolute values of the SparseSBLR parameter and the SBLR  parameter on the birds dataset in Figure \ref{heatmap_sparsesblr} and Figure \ref{heatmap_sblr}. In order to see what kind of class dependency is portrayed, we show the learned parameters $\{\beta_1^k, \cdots, \beta_m^k\}$ more detailedly on the birds dataset in Table \ref{birds_d}, where the parameters reflect the contributions from other classes to class $k$ (class 'Pacific Wren'). As can be seen from Table \ref{birds_d}, the contributions from the low co-occurrence classes often have smaller parameter values. For example, the contributions, learned by SparseSBLR, from three classes 'Dark-eyed Junco', 'Hammond's Flycatcher' and 'Swainson's Thrush' to class 'Pacific Wren' are $|0| < |-1.07| < |3.23|$, while their corresponding counts of co-occurrence are $0 < 13 < 18$.


\section{Conclusion}
In this paper, we propose a novel framework for MLC and MCC by uniting SML and generalized linear models. In particular, by uniting SML and logistic regression model, we present two novel approaches, called SBLR and SparseSBLR respectively. Both methods can capture interdependencies between classes and moreover, SparseSBLR can also discovery noisy classes to prevent them from impairing classification performance, thus providing the interpretability to some extent. What's more, extensive experiments on MLC and MCC datasets not only show excellent classification performance, but also reveal what kind of class dependency is portrayed by our model.

\section*{Acknowledgments}
This work is supported by the National Natural Science Foundation of China (Nos. 62076124 and 62006098) and the China Postdoctoral Science Foundation (No. 2020M681515). It is completed in the Nanjing University of Aeronautics and Astronautics.





%

\bibliographystyle{ieeetr}
\bibliography{mybibfile}

\end{document}